\documentclass[final]{cvpr}

\usepackage{times}
\usepackage{epsfig}
\usepackage{graphicx}
\usepackage{amsmath}
\usepackage{amssymb}
\usepackage{algpseudocode}
\usepackage{algorithm}
\usepackage{indentfirst}
\usepackage{tabularx,booktabs, etoolbox}
\usepackage{xcolor}
\usepackage{pifont}
\usepackage{multirow}
\usepackage{subfigure}
\usepackage{caption}
\usepackage{gensymb}

\newlength{\toprulewidth}
\patchcmd{\toprule}
  {\heavyrulewidth}{\toprulewidth}
  {}{}

\newlength{\bottomrulewidth}
\patchcmd{\bottomrule}
  {\heavyrulewidth}{\bottomrulewidth}
  {}{}
  
\usepackage[pagebackref=true,breaklinks=true,colorlinks,bookmarks=false]{hyperref}

\def\cvprPaperID{3200} 
\def\confYear{CVPR 2021}

\begin{document}



\title{Regularization Strategy for Point Cloud via Rigidly Mixed Sample}
\author{Dogyoon Lee$^{1}$ \quad
       Jaeha Lee$^{1}$ \quad
       Junhyeop Lee$^{1}$ \quad
       Hyeongmin Lee$^{1}$ \quad
       Minhyeok Lee$^{1}$ \\
       Sungmin Woo$^{1}$ \quad
       Sangyoun Lee$^{1*}$ \\ 
       \vspace{0.01cm}\\
       $^{1}$Yonsei University \\
       {\tt\small \{nemotio,jaeha0725,jun.lee,minimonia,hydragon516,smw3250,syleee\}@yonsei.ac.kr}
    }

\maketitle
\begin{abstract}
Data augmentation is an effective regularization strategy to alleviate the overfitting, which is an inherent drawback of the deep neural networks. However, data augmentation is rarely considered for point cloud processing despite many studies proposing various augmentation methods for image data. Actually, regularization is essential for point clouds since lack of generality is more likely to occur in point cloud due to small datasets. This paper proposes a Rigid Subset Mix~(RSMix)\footnote{Project page: \textit{\urlstyle{sf}\url{https://github.com/dogyoonlee/RSMix}}}, a novel data augmentation method for point clouds that generates a virtual mixed sample by replacing part of the sample with shape-preserved subsets from another sample. RSMix preserves structural information of the point cloud sample by extracting subsets from each sample without deformation using a neighboring function. The neighboring function was carefully designed considering unique properties of point cloud, unordered structure and non-grid. Experiments verified that RSMix successfully regularized the deep neural networks with remarkable improvement for shape classification. We also analyzed various combinations of data augmentations including RSMix with single and multi-view evaluations, based on abundant ablation studies.
\end{abstract}
\vspace{-0.3cm}
\begin{figure}[t!]
         \includegraphics[scale=1.0]{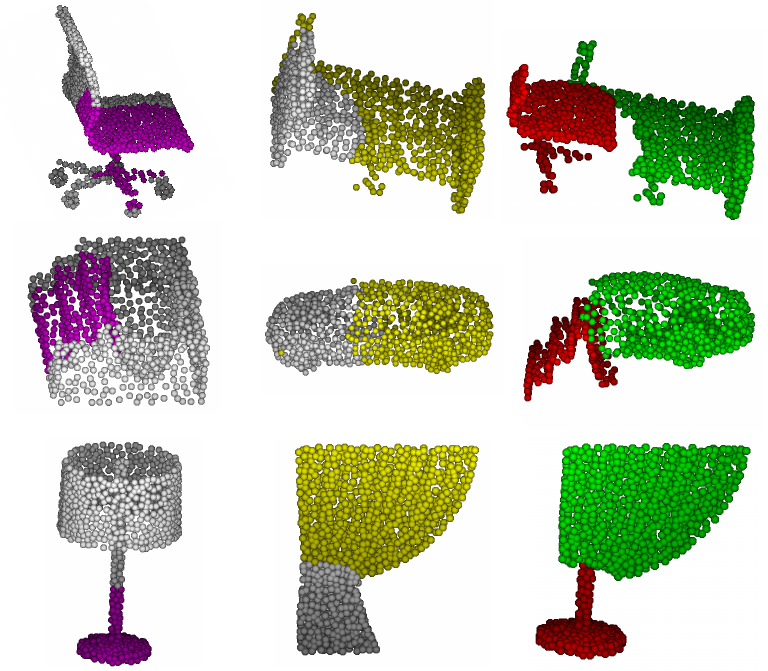} 
      \caption{Qualitative results with RSMix. Purple~(left) and yellow~(middle) colored points indicate Rigid Subsets to be extracted from each sample to synthesize red and green colored mixed samples~(right).}
	  \vspace{-0.5cm}
      \label{fig:representative_figure}
\end{figure}

\section{Introduction} \label{section:Introduction}
Deep neural networks have achieved outstanding performances in various fields regardless of the data domains, such as image, video, speech, and point cloud. In particular, three-dimensional~(3D) point cloud processing is attracting considerable interest following the pioneering network PointNet~\cite{qi2017pointnet} development, since point clouds can be applied directly to deep learning without preprocessing. Although various tasks have been successfully addressed using point clouds, such as 3D object shape classification and part segmentation, inherent drawback of deep learning is still less considered in the point cloud domain.
Due to the typical nature of deep neural networks~(DNNs) that approximates the model from the given data distribution, the trained model tends to be overfitted regardless of the data domain. This lack of generality is a fundamental deep learning problem. One way to alleviate overfitting and generalize the model is data augmentation, which improves diversity of the training data.\\
\indent Various data augmentation methods have been recently proposed in the image domain as network regularization strategies, but data augmentation for point clouds has only rarely been considered. Actually, regularization is essential for point clouds since it is easier to be biased to the distribution of training samples than that of image. That is largely because point cloud datasets~\cite{wu20153d,chang2015shapenet,dai2017scannet} are typically considerably smaller and less diverse than image datasets, such as ImageNet~\cite{deng2009imagenet} and MSCOCO~\cite{lin2014microsoft}, which have millions of training data. For example, ModelNet40~\cite{wu20153d}, one of the most widely used point cloud dataset, includes only 12,311 models with 40 categories. Therefore, it is essential to improve the generality of models for point cloud.\\
\indent In the image domain, regional dropout~\cite{zhong2020random,singh2017hide,ghiasi2018dropblock,devries2017improved} and mixup-based methods~\cite{zhang2017mixup,verma2019manifold,yun2019cutmix,kim2020puzzle,harris2020fmix} have been proposed as data augmentation strategies, which are different to conventional methods, to generate virtual training samples. These methods are designed to improve generality of the neural network, preventing the model from being significantly affected by only small part of the sample that has discriminative characteristics by eliminating or mixing the part of the data. However, it is difficult to apply this intuition directly to the point clouds due to two inherent properties of point cloud: non-grid and order-invariance. Although Chen~\etal~\cite{chen2020pointmixup} applied the concept of Mixup~\cite{zhang2017mixup} to point clouds handling the properties of point cloud with linear interpolation based on optimal assignment, generated samples lost the structural information of the original sample due to distortion.\\
\indent This paper proposes Rigid Subset Mix~(RSMix), the shape-preserving data augmentation method for point clouds that can partially mix two samples, preserving the partial shapes of the original samples. We redefine the concept of mask region from image analysis and adapt it to 3D space to extract parts from each sample while preserving structural information of the point cloud. We also define a Rigid Subset~(RS) derived from the redefined mask region, a group of adjacent points within a certain distance from a specific query point using a neighboring function to address unique characteristics: unordered structure and non-grid. In contrast to PointMixup~\cite{chen2020pointmixup}, we can utilize structural information of the original point cloud sample intactly by using RS. In addition, we designed RS scale to vary, to improve diversity of the training sample, and hence increase regularization effects. Furthermore, RSMix can be used in conjunction with the existing data augmentation since it utilizes the part of the given data intactly. In the end, by introducing RS, we can improve generality of DNNs and give attention them to recognize parts of the object. In Section~\ref{subsection:mask_region_3d}, we describe in detail how to generate the virtual sample preserving shape of the source sample by extracting RS. In advance, we provide visualized RS samples to be extracted and resultant mixed samples in \figurename~\ref{fig:representative_figure}. \\
\indent We provide the experimental results for shape classification on ModelNet40~\cite{wu20153d} and ModelNet10~\cite{wu20153d} with the most representative DNN approaches~\cite{qi2017pointnet++,wang2019dynamic} for point clouds. RSMix successfully improved the network performance, outperforming the existing data augmentation methods. Moreover, abundant ablation studies for various combinations of existing data augmentation and RSMix verified that RSMix improved the model regardless of which conventional data augmentation method was employed.\\
\indent Meanwhile, we analyzed the experimental results with respect to two evaluation mechanisms to ensure fair comparisons. In fact, although the evaluation methods of shape classification on point cloud are divided into two ways: single and multi-view, many studies present their experimental results without clearly specifying their mechanism. This makes hard to quantitatively compare results among studies. Our experiments show that the results evaluated by single and multi-view approaches have significant differences. Therefore, it is essential to analyze experimental results along the evaluation methods. Sections \ref{subsection:ablation_study} presents analysis with single and multi-view approaches based on ablation studies.\\
\indent To summarize, this paper provides the following major contributions.
\begin{itemize}
	\item \textit{Shape preserving augmentation.} We propose new data augmentation method for point clouds that mixes training samples with preserved structures by using Rigid Subset~(RS). 
	\item \textit{Significant improvement.} The proposed method remarkably improves DNN performances and robustness for shape classification and outperforms existing data augmentation strategies.
\item \textit{Complementary method.} RSMix can be used in conjunction with other data augmentation approaches. Abundant ablation studies verify that RSMix can be combined well with other augmentations and further improves the target model.
\end{itemize}

\section{Related Work} \label{section:Related_Work}
\noindent \textbf{Data Augmentation for Images.} Data augmentation is a regularization methods that expands the knowledge range that can be learned from training data by transforming data while retaining the essential sample meaning. Thus, the model becomes less dependent on the specific given data. Various methods have been proposed in the image domain in addition to conventional methods, such as random rotation, flip and crop. Some works have enabled the model to learn spatially distributed representation by removing the part of the data on pixel~\cite{zhong2020random,singh2017hide,devries2017improved} or feature map~\cite{ghiasi2018dropblock} basis. Furthermore, several mixup-based methods~\cite{zhang2017mixup,verma2019manifold,yun2019cutmix,kim2020puzzle,harris2020fmix} have been proposed that generate virtual samples by combining the two samples.\\
\indent Mixup~\cite{zhang2017mixup} generates virtual training samples by linearly interpolating two images and defining the mixed area ratio as a corresponding label. By introducing the combination between data, Mixup brings out the regularization effect and shows improved performance for several tasks. After Mixup, Verma~\etal~\cite{verma2019manifold} extended Mixup by applying the concept to the feature map, and Yun~\etal~\cite{yun2019cutmix} fusioned the concept of \cite{zhang2017mixup} and \cite{devries2017improved} to improve localization and classification ability of the model. In addition, Kim~\etal~\cite{kim2020puzzle} and Harris~\etal~\cite{harris2020fmix}  utilized saliency maps~\cite{kim2020puzzle} and Fourier transform~\cite{harris2020fmix}, respectively, to use semantically representative parts of the data when generating virtual samples. However, these approaches can only be applied to image based models rather than point clouds, because they have different data structures. Therefore, we propose the RSMix, novel mixup-based augmentation strategy for the point cloud that generates virtual samples considering the unique properties of point clouds .\\[2pt]
\begin{table}
   \footnotesize
      \resizebox{\columnwidth}{!}{
         \begin{tabular}{lc}
	        \specialrule{0.6pt}{1pt}{1pt}
			Method & Mix function \(f_{M_d}(x_{\alpha}, x_{\beta})\)\\
			\midrule\midrule
            Mixup \cite{zhang2017mixup} & \((1-\lambda)x_{\alpha}+\lambda x_{\beta}\)\\
			Manifold Mixup \cite{verma2019manifold} & \((1-\lambda)f(x_{\alpha})+\lambda f(x_{\beta})\)\\
			CutMix \cite{yun2019cutmix} & \((1-\mathcal{M})\odot x_{\alpha}+\mathcal{M}\odot x_{\beta}\)\\
			Puzzle Mix \cite{kim2020puzzle} & \((1-z)\odot\Pi^{T}_{\alpha} x_{\alpha}+z\odot\Pi^{T}_{\beta}x_{\beta}\)\\
			F-Mix \cite{harris2020fmix} & \((1-\mathcal{H}(\mathcal{G}))\odot x_{\alpha}+\mathcal{H}(\mathcal{G})\odot x_{\beta}\)\\
			PointMixup \cite{chen2020pointmixup} & \((1-\lambda)x_{\alpha}+\lambda \mathcal{J}_{\phi^{*}}(x_{\alpha},x_{\beta})\)\\
	        \specialrule{0.6pt}{1pt}{1pt}
      \end{tabular}}%
   \caption{Various mixup functions for image and point cloud domains.}
   \vspace{-0.5cm}
   \label{table:mixupfunctions}
\end{table}
\noindent \textbf{Point Cloud Structural Properties.} In contrast with images, point clouds have 3D coordinate information, including implicit geometric feature, which is essential to understand them. Due to the unique properties of point cloud: non-grid and unordered structures, it is difficult to extract the local and geometric feature from point clouds. Various networks have been proposed with different structures, such as point-wise multi layer perceptron~(MLP)~\cite{qi2017pointnet++,joseph2019momen,zhao2019pointweb,yan2020pointasnl}, convolution~\cite{liu2019relation,thomas2019kpconv,wu2019pointconv,boulch2020convpoint,hua2018pointwise,lei2019octree,li2018pointcnn,lan2019modeling}, graph~\cite{wang2019dynamic,zhang2019linked,shen2018mining,zhang2018graph}, and spatial partitioning structured~\cite{gadelha2018multiresolution, klokov2017escape} based networks to extract local and geometric features. For example, ~\cite{qi2017pointnet++,thomas2019kpconv,wang2019dynamic, klokov2017escape} extract local and geometric features by applying point-wise grouping, radius-based kernel, graph-structure, and space partitioning tree, respectively. All these networks demonstrated that considering structural information is significantly important for the DNN model to understand point clouds. Therefore, regardless of how data augmentation occurs, structural information of point cloud should be regarded as core characteristics since it is a critical component for the model.\\[2pt]
\noindent \textbf{Data Augmentation on Point Cloud.} Data augmentation has not been extensively explored in the point cloud domain, aside from general conventional methods, such as randomly scaling, rotation, and jittering. Few studies~\cite{li2020pointaugment,choi2020part,chen2020pointmixup} dealt with data augmentation in the point cloud domain. Liu~\etal\cite{li2020pointaugment} proposed auto-augmentation network for point clouds to find an optimal combination of conventional data augmentation methods corresponding to each sample. Choi~\etal\cite{choi2020part} divided the sample into specific partitions and transformed or mixed each part independently. However, there is a limit to diversity of virtual sample because mixing is performed on inter-classes and specified grids are used for partitioning. PointMixup~\cite{chen2020pointmixup}, which is the closest method to our proposed approach, extends the concept of Mixup~\cite{zhang2017mixup} to point clouds through linear interpolation with optimal assignments between two samples. However, the generated samples have distorted structures which lead to the loss of structural information. Structural information is a core point cloud feature since they have no textural information. Therefore, we propose a more general data augmentation method for point clouds that can preserve shape of the original data .

\begin{figure*}
      \captionsetup{justification=centering}
         \includegraphics[scale=1.0]{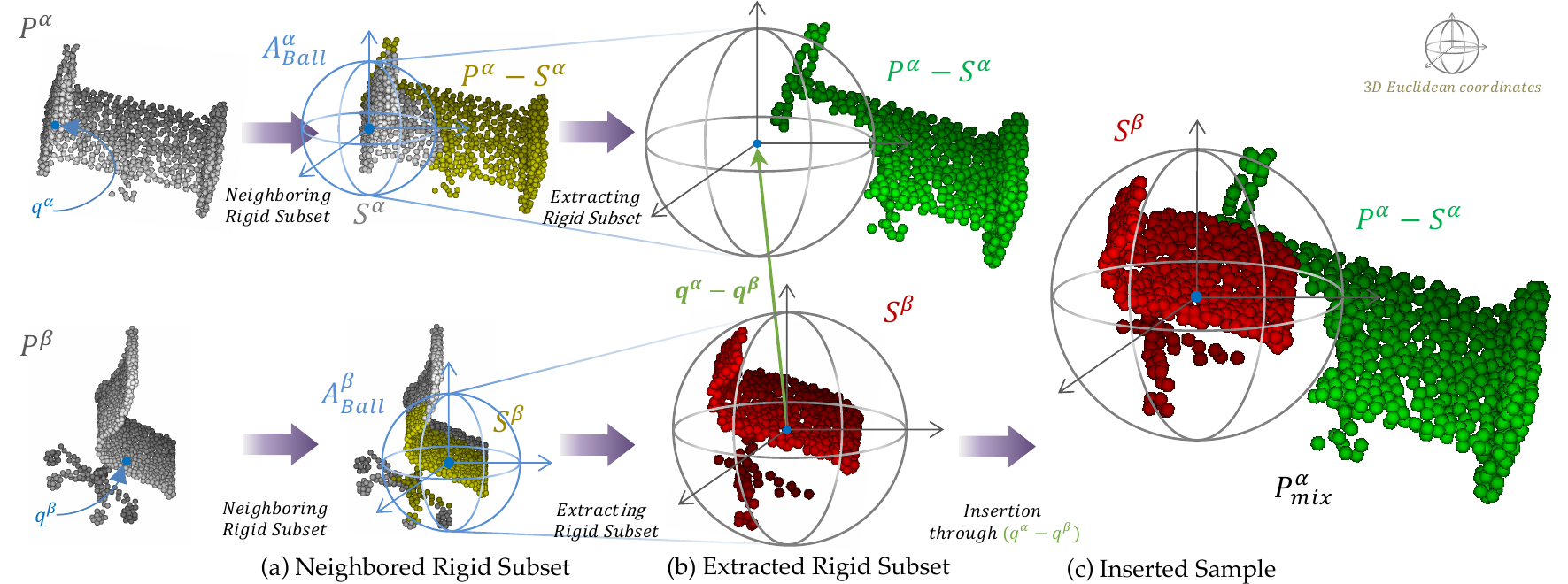} 
      \caption{Overall pipeline of RSMix. Three steps to synthesize the mixed samples(\(\mathcal{P}^{\alpha}_{mix}\)) using Rigid Subset~(RS).}
      \vspace{-0.3cm}
      \label{fig:RSMix_pipeline}
   \end{figure*}

\vspace{-0.2cm}
\section{Method} \label{section:method}
\subsection{Preliminary}\label{subsection:Preliminary}
Neural networks aim to model function \(f\) that describes the true distribution \(P\) for given data \(\mathcal{D}=\{(x_i,y_i)\}^n_{i=1}\), where samples \(x\in\mathcal{X}\) have corresponding labels \(y\in\mathcal{Y}\). It has been proved 
through \textit{Empirical Risk Minimization}~\cite{vapnik1971uniform} that \(f\) can be approximated by minimizing empirical risk \(R_{\xi}(f)\) of the model by computational optimization using loss \(\mathcal{L}\) and empirical distribution \(P_{\xi}\) for given data distribution as
\vspace{-0.3cm}
\begin{equation} \label{eq:EVR}
R_{\xi}(f)=\int\mathcal{L}(f(x),y)dP_{\xi}(x,y)= \cfrac{1}{n}\sum_{i=1}^{n}\mathcal{L}(f(x_i),y_i)
\end{equation}
In data augmentation, \(P_{\xi}\) can be expanded to \(P_{\psi}\) with additional augmented data through \textit{vicinal risk minimization}~\cite{chapelle2001vicinal},
\vspace{-0.2cm}
\begin{equation} \label{eq:VRM}
\vspace{-0.2cm}
P_{\psi}(\tilde{x},\tilde{y})=\cfrac{1}{n}\sum_{i=1}^{n}\psi(\tilde{x},\tilde{y}|x_i,y_i),
\end{equation}
where \(\psi\) is a vicinity distribution, \ie, the probability that virtual sample and label pair \((\tilde{x},\tilde{y})\), are sampled from the vicinity of given sample and label pair \((x_{i},y_{i})\). For image data, Zhang~\etal~\cite{zhang2017mixup} designed a vicinal distribution \(\psi\) that generated a virtual mixed sample-label pair \((\tilde{x},\tilde{y})\) from two paired data \((x_{\alpha},y_{\alpha})\) and \((x_{\beta},y_{\beta})\) using mix function \(f_{M_{d}}\), for sample and \(f_{M_l}\), for label, as
\begin{equation} \label{eq:Mixup_mixupfunctions}
\vspace{-0.2cm}
\begin{split}
\tilde{x}=f_{M_{d}}(x_{\alpha},x_{\beta})=(1-\lambda)x_{\alpha}+\lambda x_{\beta},\\
\tilde{y}=f_{M_{l}}(y_{\alpha},y_{\beta})=(1-\lambda)y_{\alpha}+\lambda y_{\beta},
\end{split}
\end{equation}
where \(\lambda \sim\) beta distribution Beta\((\theta,\theta)\), for \(\theta\in(0,\infty)\). \tablename~\ref{table:mixupfunctions} shows the deformations of \(f_{M_d}\) in various ways using features from model \(f\)~\cite{verma2019manifold} or masking approaches such as binary mask \(\mathcal{M}\)~\cite{yun2019cutmix}, salient data included mask \(z\)~\cite{kim2020puzzle}, or thresholding mask \(\mathcal{H}\)~\cite{harris2020fmix} with filtered data \(\mathcal{G}\) in the frequency domain. However, these mask-based approaches cannot be applied directly to point cloud, since point clouds have no grid and points can exist anywhere in 3D real space. Though Chen~\etal~\cite{chen2020pointmixup} solved this problem by linear interpolation between two point clouds, introducing optimal assignment \(\mathcal{J}_{\phi^{*}}\), they could not generate virtual samples preserving shape of the original sample. Therefore, our goal is to generate a shape-preserved virtual sample that has combined information from both samples as well as proposing an adapted spatial mask for 3D data. We are inspired by concept of the mask region from image analysis, which preserves the part of the original data intactly.\\[2pt] 
\textbf{Mask as Region of Neighboring Data.} CutMix~\cite{yun2019cutmix} defined the mix function as 
\vspace{-0.1cm}
\begin{equation} \label{eq:cut_mix_mixfunction}
f_{M_d}(x_{\alpha},x_{\beta})=(1-\mathcal{M})\odot x_{\alpha}+\mathcal{M}\odot x_{\beta},
\vspace{-0.1cm}
\end{equation}
where \(\odot\) represents the element-wise multiplication; mask \(\mathcal{M}\) denotes a  \(d_u\times d_w\) binary rectangular region represented as \([u_M,u_M+d_u]\times[w_M,w_M+d_w]\) with mixture ratio, \(\lambda=\frac{d_u d_w}{WH}\), where \((u_M,w_M)\) is a randomly chosen pixel of the image. To utilize inherent definition of the mask region, we redefine mask \(\mathcal{M}\) as a group of successive adjacent pixels within distances \(d_u/2\) and \(d_w/2\) in image. Hence, mask \(\mathcal{M}\) can be denoted as 
\vspace{-0.1cm}
\begin{equation} \label{eq:mask_defi}
\mathcal{M} = \{(u_{i},w_{j})\vert\;\vert u_i - u_c\vert \leq \frac{d_u}{2},\vert w_j - w_c\vert \leq \frac{d_w}{2}\,\},
\vspace{-0.2cm}
\end{equation}
where \((u_i,w_j)\) is the \((i,j)th\) pixel for the given image; \(i=\{1,2,...,W\}\); \(j=\{1,2,...,H\}\); and \((u_c,w_c)\) is the center of the mask \((u_M+\frac{d_u}{2}, w_M+\frac{d_w}{2})\). Thus, the mask can be regarded as an adjacent group of data from a particular point \((u_{c},w_{c})\). We adapt this definition of mask to point clouds.

\subsection{Rigid Subset Mix} \label{subsection:mask_region_3d}
Rigid Subset Mix~(RSMix) mixes parts of two point cloud samples by extracting the Rigid Subset~(RS), which preserves each samples's shape. RSMix is divided in three steps: neighboring, extraction, and insertion. First, we utilize the redefined mask region concept from Section~\ref{subsection:Preliminary} with neighboring function \(\mathcal{A}\) to prevent deformation of original data. Then we extract RSs from each sample to mix the samples. Finally, we mix two RSs in the insertion step. \figurename~\ref{fig:RSMix_pipeline} shows the RSMix algorithm pipeline.\\[2pt]
\noindent \textbf{Neighboring Rigid Subset.} We define two \(n\) sampled point sets normalized in the unit sphere as \(\mathcal{P}^t=\{p^{t}_i\,\vert\; i=1,2,...,n\}\), where t\(\in\{\alpha,\beta\}\). \(p^{t}_{i}\in \mathbb{R}^{3}\) is its Euclidean coordinates, which represents location of the point. We only consider coordinate information since RSMix operates on point-wise coordinates.\\
\indent We adapt the regional mask for image to spatial subset of each point sets from given point sets \(\mathcal{P}^\alpha\) and \(\mathcal{P}^\beta\), by grouping adjacent points from a certain query point, \(q^{t}\), randomly chosen from \(\mathcal{P}^t\). These subsets are denoted as \(\mathcal{S}^{\alpha}\) and \(\mathcal{S}^{\beta}\) according to below Equation (\ref{eq:Rigid_Subset_Mask}).
\begin{equation} \label{eq:Rigid_Subset_Mask}
\mathcal{S}^{t}=\mathcal{A}(\mathcal{P}^t{;}q^t),
\end{equation}
which are grouped using the specific neighboring function \(\mathcal{A}\). We define these subsets as Rigid Subset~(RS) since they preserve the sample shape rigidly. \\
\indent We introduce two instantiations for \(\mathcal{A}\) to retain the original point set shape: K-Nearest Neighbor(KNN) to given \(q^{t}\) and Ball-query method that neighboring points in certain distance \(r_{rigid}\) from \(q^{t}\) as 
\begin{equation} \label{eq:knn_ball-query_function}
\begin{split}
\mathcal{A}_{knn}(\mathcal{P}^t{;}q^t) & =\{\,p^t\vert\;p^{t}\;\text{is KNN of}\;q^t\,,p^t\in\mathcal{P}^{t}\},\\
\mathcal{A}_{ball}(\mathcal{P}^t{;}q^t) & =\{\,p^t\,\vert\;\Vert p^t-q^t\Vert \leq r_{rigid}\,,p^t\in\mathcal{P}^{t}\},
\end{split}
\end{equation}
respectively, where \(r_{rigid}\) is sampled from beta distribution Beta\((\theta,\theta)\), with parameter \(\theta=1.0\) as default, \ie, the uniform distribution, since \(P^t\) is normalized in the unit sphere. Both \(\mathcal{A}\) are based on Euclidean distance in 3D space considering  the point cloud's unordered structure and free space around them. Each method has different characteristics on neighboring subsets with respect to the density or directional bias of given point sets. \\
\indent Meanwhile, we limit \(|\mathcal{S}^{\beta}|\) \(\leq\) \(n^{max}\), where \(n^{max}\) and \(|\cdot|\) denote the upper bound number of points in RS and cardinality for the point set, respectively. We usually set \(n^{max}=|\mathcal{P}^{t}|/2\) to preserve at least half of the original point sets. In addition, when using \(\mathcal{A}_{ball}\), we randomly sample points in \(\mathcal{S}^{\beta}\) along the difference between \(|\mathcal{S}^{\alpha}|\) and \(|\mathcal{S}^{\beta}|\) to maintain the \(|\mathcal{P}^{\alpha}_{mix}|\), where \(\mathcal{P}^{\alpha}_{mix}\) denotes a mixed sample described in below Insertion part. We compare and analyze the two methods for quantitative and visualized results in Section \ref{subsection:ablation_study}. Further, we also provide experiments with various \(\theta\) values in Section \ref{subsection:ablation_study}. \figurename~\ref{fig:RSMix_pipeline}(a) shows neighboring the RS from the each sample.\\[2pt]
\noindent \textbf{Extraction.} Neighbored RSs, \(\mathcal{S}^{\alpha}\) and \(\mathcal{S}^{\beta}\), are used to a generate mixture sample \(\mathcal{P}^{\alpha}_{mix}\). To mix two samples, we remove the \(\mathcal{S}^{\alpha}\) from \(\mathcal{P}^{\alpha}\) and replace the empty space with \(\mathcal{S}^{\beta}\). Hence, extracted RSs from each point cloud sample to generate mixture samples are denoted as \(\mathcal{P}^{\alpha}-\mathcal{S}^{\alpha}\) and \(\mathcal{S}^{\beta}\) as shown in \figurename~\ref{fig:RSMix_pipeline}(b).\\[2pt]
\noindent \textbf{Insertion.} However, \(q^{\alpha}\) and \(q^{\beta}\) are usually different because they are randomly chosen from \(\mathcal{P}^{\alpha}\) and \(\mathcal{P}^{\beta}\), respectively. Hence, before insertion, \(\mathcal{S}^{\beta}\) should be translated by the difference between the  two query points. We introduce translation function \(\mathcal{T}^{\beta\rightarrow\alpha}\) to translate \(\mathcal{S}^{\beta}\) by \(q^{\alpha}-q^{\beta}\) as 
\begin{equation} \label{eq:translation_function}
\mathcal{T}^{\beta\rightarrow \alpha}(\mathcal{S}^{\beta};q^{\alpha},q^{\beta}) = \{p^{\beta\rightarrow\alpha}\,\vert\;p^{\beta\rightarrow\alpha}=p^{\beta}+(q^{\alpha}-q^{\beta})\},
\end{equation}
where \(p^{\beta}\) is a point in \(\mathcal{S}^{\beta}\). Applying \(\mathcal{T}^{\beta\rightarrow\alpha}\) to \(\mathcal{S}^{\beta}\), the translated subset \(\mathcal{S}^{\beta\rightarrow \alpha}\) is denoted as
\begin{equation} \label{eq:traslated_subset_mask}
\begin{split}
\mathcal{S}^{\beta\rightarrow \alpha} = \mathcal{T}^{\beta\rightarrow \alpha}(\mathcal{S}^{\beta};q^{\alpha},q^{\beta}).
\end{split}
\end{equation}
Therefore, mixed sample \(\mathcal{P}^{\alpha}_{mix}\) is defined as 
\begin{equation} \label{eq:mixed_sample}
\mathcal{P}_{mix}^{\alpha} = (\mathcal{P}^{\alpha}-\mathcal{S}^{\alpha})\;\cup\;\mathcal{S}^{\beta\rightarrow \alpha}\;,
\end{equation}
and \figurename~\ref{fig:RSMix_pipeline}(c) describes the inserted mixture sample. Thus, mix function \(f_{M_d}(x_{\alpha},x_{\beta})\) for RSMix can be expressed using as follow Equation~(\ref{eq:final_mix_function}) using \(\mathcal{P}^{\alpha}\) and \(\mathcal{P}^{\beta}\) instead of \(x_{\alpha}\) and \(x_{\beta}\).
\begin{equation} \label{eq:final_mix_function}
f_{M_d}(\mathcal{P}^{\alpha},\mathcal{P}^{\beta})=(\mathcal{P}^{\alpha}-\mathcal{A}(\mathcal{P}^{\alpha}))\cup \mathcal{T}^{\beta\rightarrow\alpha}(\mathcal{A}(\mathcal{P}^{\beta})),
\end{equation}
where input arguments related to query points \(q^{\alpha}\) and \(q^{\beta}\) are omitted for clarity.

\subsection{Mixture Ratio \(\lambda\) for Training} \label{subsection:mixture_ratio}
In this Section, we define the mixture ratio \(\lambda\), the ratio of \(|\mathcal{S}^{\beta\rightarrow\alpha}|\) \wrt \(|\mathcal{P}_{mix}^{\alpha}|\), to train the network for shape classification. In contrast to \(\mathcal{A}_{knn}\) or some previous image masks, \(|\mathcal{P}^{\alpha}-\mathcal{S}^{\alpha}|\) and \(|\mathcal{S}^{\beta}|\) are often different when using the \(\mathcal{A}_{ball}\), since we apply same \(r_{rigid}\) to \(\mathcal{P}^{\alpha}\) and \(\mathcal{P}^{\beta}\), despite of their different densities. Hence, we define
\begin{equation}\label{eq:lambda_cases}
\footnotesize
\lambda=\begin{cases}
			\hfil 0\;, & \text{if $\mathcal{P}^{\alpha}=\mathcal{S}^{\alpha}$,}\\
			\hfil 0\;, & \text{if $\mathcal{S}^{\beta} = \emptyset$,}\\
			\hfill |\mathcal{S}^{\beta}|\;/\;(|\mathcal{P}^{\alpha}-\mathcal{S}^{\alpha}|+|\mathcal{S}^{\beta}|)\;, & \text{Otherwise,}
		 \end{cases}
\end{equation}
To explicitly consider the relation between \(|\mathcal{P}^{\alpha}-\mathcal{S}^{\alpha}|\) and \(|\mathcal{S}^{\beta}|\). Finally, we define the label mix function as
\begin{equation} \label{eq:label_mix_function}
f_{M_l}(y_{\alpha},y_{\beta})=(1-\lambda)y_{\alpha}+\lambda y_{\beta},
\end{equation}
which is same as in CutMix~\cite{yun2019cutmix}, to generate virtual label \(\tilde{y}\) for classification training. Detailed implementations are available in the supplementary material with pseudo-code.

\section{Experiments}
\vspace{-0.1cm}
\noindent \textbf{Datasets.} We evaluate RSMix on ModelNet40~\cite{wu20153d} and ModelNet10~\cite{wu20153d}, which are widely used point cloud classification benchmark datasets. ModelNet40 is small dataset which comprises 12,311 CAD models from 40 man-made object categories, and ModelNet10 is subset of ModelNet40 that includes only 4899 CAD models from 10 categories. We utilized the preprocessed data provided by PointNet~\cite{qi2017pointnet} for ModelNet40 with same train-test split, which were 1024 uniformly sampled points on mesh faces according to face area and normalized onto the unit sphere, and preprocessed ModelNet10 similarly. In particular, we ignored normals of samples since they are not available for real-data.\\[2pt]
%
\noindent \textbf{Backbone Networks.} We considered three representative point-wise DNNs for point cloud: PointNet++~\cite{qi2017pointnet++}, DGCNN~\cite{wang2019dynamic}, and PointNet~\cite{qi2017pointnet} as our backbone network architecture. We applied RSMix to several neural networks to emphasize RSMix is model agnostic.\\[2pt]
\noindent \textbf{Single and Multi-view Evaluations.} Single and multi-view evaluations are separated depending on whether objects were evaluated from different angles or not. These approaches can be separated into two cases: with or without voting strategy to predict an object multiple times by rotating about an axis. The experiments adopted voting strategy of evaluating an object 12 times, rotating it 30\(\degree\) on its vertical~(y) axis between evaluations. Meanwhile, ModelNet40~\cite{wu20153d} has 10 classes with aligned poses/headings. Thus, it is trivial to separate the 10 classes with the remaining 30 classes if we don't do random rotation on test samples when evaluating the model. Hence, there are obvious differences between the results from single and multi-view evaluations. Appropriate combinations of augmentation strategies also vary depending on evaluation type. We investigated results from both evaluation strategies and explored optimal combinations of augmentations for different models by abundant ablation studies~(Sections~\ref{subsection:ablation_study}).\\[2pt]
\noindent \textbf{Implementation details.} We implemented RSMix using PointNet++~\cite{qi2017pointnet++} and DGCNN~\cite{wang2019dynamic} with conventional data augmentation, ConvDA~(comprising jittering(\(\sigma^{2}\)=0.01); scaling(0.8\(\sim\)1.25); rotation along the y-axis \ie, gravity axis; and shifting~(range=0.1) for the training dataset. Further details are included in the supplementary material.

\begin{table}
      \resizebox{\columnwidth}{!}{
         \begin{tabular}{lcccc}
            \toprule
            \multirow{2}{*}{Method} & \multirow{2}{*}{\#Points} & \multicolumn{2}{c}{Evaluation Accuracy(\%)} \\
            & & ModelNet40 & ModelNet10 \\
			\midrule\midrule
            PointNet\cite{qi2017pointnet} & 1k  & 88.5 & 93.1  \\
            PointNet\cite{qi2017pointnet} (Multi) & 1k  & 88.4  & 92.5  \\
            PointNet{+}{+}\cite{qi2017pointnet++}  & 1k  & 91.0  & 93.3  \\
            PointNet{+}{+}\cite{qi2017pointnet++} (Multi)  & 1k  & 91.0  & 93.5  \\
            DGCNN\cite{wang2019dynamic} & 1k & 92.8 & 94.8 \\
            \midrule
            PointNet \textbf{+Ours} & 1k  & \textbf{88.7(0.2\(\uparrow\))}  & 93.1(-)  \\
            PointNet \textbf{+Ours} (Multi) & 1k  & \textbf{88.5(0.1\(\uparrow\))}  & \textbf{92.6(0.1\(\uparrow\))} \\
            PointNet{+}{+} \textbf{+Ours}  & 1k  & \textbf{91.6(0.6\(\uparrow\))}  & \textbf{94.3(1.0\(\uparrow\))}  \\
            PointNet{+}{+} \textbf{+Ours} (Multi)  & 1k  & \textbf{92.1(1.1\(\uparrow\))}  & \textbf{94.4(0.9\(\uparrow\))}  \\
            DGCNN \textbf{+Ours} & 1k & \textbf{93.5(0.7\(\uparrow\))} & \textbf{95.9(1.1\(\uparrow\))} \\
            \bottomrule
      \end{tabular}}%
   \caption{Quantitative results for single and Multi-view evaluations of RSMix on ModelNet40~\cite{wu20153d}. We only present the results of PointNet~\cite{qi2017pointnet} and PointNet++~\cite{qi2017pointnet++} with rotational augmentation included model for fair comparison.}
      \label{table:Evaluation_results}
\end{table}
\begin{table}
      \resizebox{\columnwidth}{!}{
         \begin{tabular}{lcccc}
            \toprule
            Method & Augmentation & ACC.(\%) & Dataset Align & Eval  \\
			\midrule\midrule
            PointNet{+}{+}~\cite{qi2017pointnet++} & PointMixup~\cite{chen2020pointmixup} & \textbf{92.7}  & Pre-aligned  & -  \\
            DGCNN~\cite{wang2019dynamic} & PointMixup~\cite{chen2020pointmixup} & 93.1 & Pre-aligned & - \\
            PointNet~\cite{qi2017pointnet} & PointMixup~\cite{chen2020pointmixup} & \textbf{89.9}  & Unaligned & - \\
            PointNet{+}{+}~\cite{qi2017pointnet++} & PointMixup~\cite{chen2020pointmixup}  & 91.7  & Unaligned & -\\
            \midrule
            PointNet{+}{+}~\cite{qi2017pointnet++} & Ours & \textbf{92.7}  & Raw  & Single-View  \\
            DGCNN~\cite{wang2019dynamic} & Ours & \textbf{93.5} & Raw & Single-View  \\
            PointNet~\cite{qi2017pointnet} & Ours & 88.5 & Raw & Multi-View \\
            PointNet{+}{+}~\cite{qi2017pointnet++} & Ours & \textbf{92.1}  & Raw & Multi-View  \\
            \bottomrule
      \end{tabular}}%
   \caption{Comparing RSMix and PointMixup~\cite{chen2020pointmixup} on ModelNet40~\cite{wu20153d}.}
   \vspace{-0.4cm}
   \label{table:Comparison_PointMixup}
\end{table}

\subsection{Shape Classification} \label{subsection:shape_classification}
\vspace{-0.1cm}
\noindent \textbf{Evaluations.} We evaluate RSMix for shape classification using three backbone networks on ModelNet40 and ModelNet10. All experiments were implemented using official codes and results are shown in \tablename~\ref{table:Evaluation_results}. "Multi" indicates the evaluation with multi-view. To ensure fair comparison given the rotational bias in ModelNet40, we exclude experimental results of point-wise MLP networks~\cite{qi2017pointnet,qi2017pointnet++} trained without rotational augmentation in \tablename~\ref{table:Evaluation_results}. Section \ref{subsection:ablation_study} presents a rotation-related ablation study.\\
\indent All the results reveal that RSMix improved the network accuracies regardless of network type or evaluation methods, verifying the effectiveness of our shape-preserved mixture approach with significant improvements for PointNet++~\cite{qi2017pointnet++} and DGCNN~\cite{wang2019dynamic}, which encode local or geometric features of object through hierarchical grouping or graph structure, respectively.\\[2pt]
\noindent \textbf{Comparison against PointMixup~\cite{chen2020pointmixup}.} We demonstrate results of RSMix for two evaluation methods against PointMixup~\cite{chen2020pointmixup}, the closest work to us, in \tablename~\ref{table:Comparison_PointMixup}. We compared pre-aligned and unaligned settings to for single and multi-view accuracies, respectively, since PointMixup~\cite{chen2020pointmixup} do not specify their evaluation method but each are similar. They follow the PointCNN~\cite{li2018pointcnn} setting discriminating pre-aligned and unaligned with horizontal rotation on point cloud samples. They randomly rotate the training point cloud along the y-axis for unaligned settings. For natural evaluation, we do not preprocess the dataset as pre-aligned or unaligned~(denoted as Raw in \tablename~\ref{table:Comparison_PointMixup}). RSMix achieves more competitive performance than PointMixup~\cite{chen2020pointmixup} for networks that use local information ~\cite{qi2017pointnet++,wang2019dynamic}, and further enhances the network’s ability to recognize local information.\\[2pt]
\noindent \textbf{Visualization.} Supplementary material provides additional examples synthesized with RSMix. 

\begin{table}
   \footnotesize
      \resizebox{\columnwidth}{!}{
         \begin{tabular}{ccccc}
	        \specialrule{0.7pt}{1pt}{1pt}
            ConvDA & RandDrop  & RSMix & ACC\(^{knn}_{S}\)(\%) & ACC\(^{ball}_{S}\)(\%)  \\
			\midrule\midrule
             &  & \ding{51} & 93.0(0.5\(\uparrow\)) & \textbf{93.3(0.8\(\uparrow\))}   \\
            \ding{51} &  & \ding{51} & 93.3(0.7\(\uparrow\)) & \textbf{93.4(0.8\(\uparrow\))}  \\
            \ding{51} & \ding{51} & \ding{51} & 93.4(0.6\(\uparrow\))  & \textbf{93.5(0.7\(\uparrow\))}  \\
	        \specialrule{0.7pt}{1pt}{1pt}
      \end{tabular}}%
   \caption{Quantitative Comparison of neighboring functions for DGCNN~\cite{wang2019dynamic} on ModelNet40. ACC\(^{knn}_{S}\) and ACC\(^{ball}_{S}\) indicate single-view accuracy with \(\mathcal{A}_{knn}\) and \(\mathcal{A}_{ball}\).}
   \vspace{0.4cm}
   	\label{table:knn_vs_ball_query}
\end{table}
\begin{figure}
      \captionsetup{justification=centering}
         \includegraphics[scale=1.0]{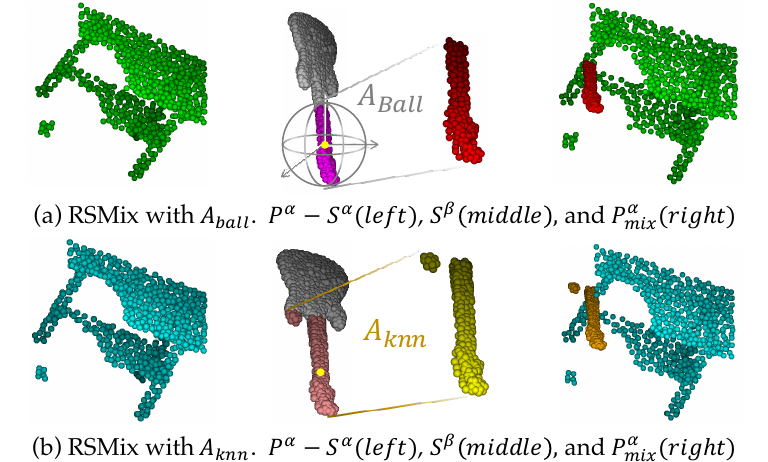} 
      \caption{Differences depending on directional\\bias of point cloud.}
      \label{fig:knn_ball_query_2}
      \vspace{0.3cm}
\end{figure}
\begin{figure}
      \captionsetup{justification=centering}
         \includegraphics[scale=1.0]{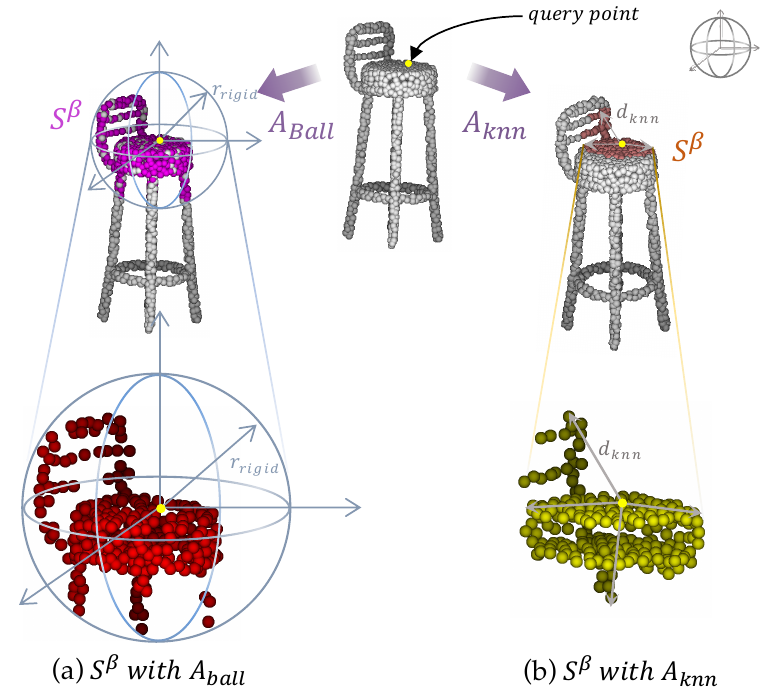} 
      \caption{Differences depending on density of point cloud. }
      \vspace{-0.3cm}
      \label{fig:knn_ball_query}
\end{figure}

\subsection{Ablation Study} \label{subsection:ablation_study}
\begin{figure}[t!]
      \captionsetup{justification=centering}
         \includegraphics[scale=1.0]{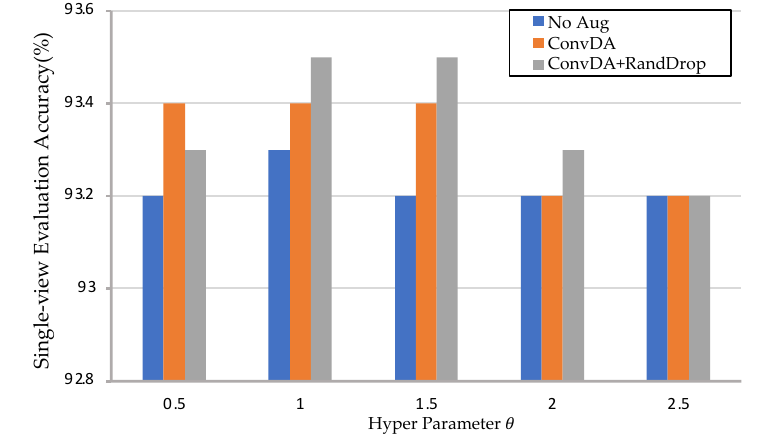} 
      \vspace{-0.1cm}
      \caption{Ablation studies for various \(\theta\) values on DGCNN~\cite{wang2019dynamic} with single-view evaluation on ModelNet40.}
      \label{fig:alpha_test}
      \vspace{-0.4cm}
\end{figure}

\noindent \textbf{Neighboring Function.} This Section describes the two types of neighboring function, \(\mathcal{A}_{ball}\) and \(\mathcal{A}_{knn}\), employed to extract RSs from point cloud samples. \figurename~\ref{fig:knn_ball_query_2} and \ref{fig:knn_ball_query} show the differences between these functions under directional bias or different densities between point cloud samples qualitatively. Although \(\mathcal{A}_{knn}\) extracts the subset preserving shape of the sample based on Euclidean distance \(d_{knn}\), it is prone to be overlapped with the other extracted part, \eg \figurename~\ref{fig:knn_ball_query_2}(b), if there is directional bias on the sample. On the other hand, \(\mathcal{A}_{ball}\) alleviates overlapping by selecting points within the distance, \(r_{rigid}\) \eg \figurename~\ref{fig:knn_ball_query_2}(a). However, in contrast with \(\mathcal{A}_{knn}\), if the density of sample around the query point is high or \(r_{rigid}\) is too large, the number of points in RS from \(\mathcal{A}_{ball}\) must be controlled to maintain the number of points in a mixed sample. Therefore, we randomly sample the extracted points in \(\mathcal{S}^{\beta}\) to preserve the overall shape of the extracted part. \figurename~\ref{fig:knn_ball_query} illustrates the difference between processed RS with \(\mathcal{A}_{ball}\) and \(\mathcal{A}_{knn}\) depending on density of point cloud. We also compare two methods with quantitative results of single-view evaluation on DGCNN~\cite{wang2019dynamic} in \tablename~\ref{table:knn_vs_ball_query}, using scaling and shift augmentations as ConvDA. Both approaches achieved improved results over existing networks and \(\mathcal{A}_{ball}\) achieves superior results since it is more robust to directional bias as well as considers the density of point cloud. Therefore, we used \(\mathcal{A}_{ball}\) as our main neighboring function.\\[2pt]
%
\begin{table}
      \resizebox{\columnwidth}{!}{
         \begin{tabular}{ccccccc}
            \toprule
            Jitter+Shift & Rotation & Scaling & RandDrop & RSMix & ACC\(_{S}\) & ACC\(_{M}\) \\
			\midrule\midrule
			& & & & & 91.5 & 78.5 \\
			& & & & \ding{51} & \textbf{92.7(1.2\(\uparrow\))} & 71.5(7.0\(\downarrow\)) \\
			\midrule
			\ding{51} &  &  & & & 91.4 & 73.5 \\
			\ding{51} &  &  & & \ding{51} & 92.0(0.6\(\uparrow\)) & 74.4(1.1\(\uparrow\)) \\
			\ding{51} &  &  & \ding{51} & & 91.4 & 67.1 \\
			\ding{51} &  &  & \ding{51} & \ding{51} & 91.8(0.4\(\uparrow\)) & 72.8(5.7\(\uparrow\)) \\
			\midrule
			& \ding{51} & \ding{51} & & & 91.0 & 90.8 \\
			& \ding{51} & \ding{51} & & \ding{51} & 91.6(0.6\(\uparrow\)) & \textbf{92.1(1.3\(\uparrow\))} \\
			& \ding{51} & \ding{51} & \ding{51} & & 91.0 & 91.0 \\
			& \ding{51} & \ding{51} & \ding{51} & \ding{51} & 91.3(0.3\(\uparrow\)) & 91.2(0.2\(\uparrow\)) \\
			\midrule
			\ding{51} & \ding{51} & \ding{51} & & & 90.3 & 90.7 \\
			\ding{51} & \ding{51} & \ding{51} & & \ding{51} & 90.8(0.5\(\uparrow\)) & 91.4(0.7\(\uparrow\)) \\
			\ding{51} & \ding{51} & \ding{51} & \ding{51} & & 90.6 & 90.7 \\
			\ding{51} & \ding{51} & \ding{51} & \ding{51} & \ding{51} & 91.0(0.4\(\uparrow\)) & 91.1(0.4\(\uparrow\)) \\
            \bottomrule
      \end{tabular}}%
   \caption{Ablation studies on evaluation accuracy with single(ACC\(_{S}\)(\%)) and multi-view(ACC\(_{M}\)(\%)) for PointNet++~\cite{qi2017pointnet++} on ModelNet40.}
	\vspace{-0.4cm}
      \label{table:pointnet2_ablation}
\end{table}
\noindent \textbf{Single and multi-view evaluations with various combinations of augmentations.} RSMix can be applied in conjunction with existing ConvDA methods to further increase the diversity of mixed data since they are independent approaches. However, some combinations of augmentations can cause excessive deformation on the data sample, reducing the network's ability to recognize objects itself. Therefore, it is essential to analyze various combinations of augmentation strategies. We provide quantitative results in \tablename~\ref{table:pointnet2_ablation} with single and multi-view evaluations for PointNet++~\cite{qi2017pointnet++} on ModelNet40. RandDrop is the data augmentation method proposed in \cite{qi2017pointnet++} that randomly drops the points from sample so that network can extract the global features better. The results show that models with RSMix alone achieved the highest accuracy for single-view evaluation. In addition, overall experiments show better results without rotational augmentation for single-view evaluation. However, results with multi-view evaluation reveal that if the model is trained without rotational augmentation, network can be overfitted to directional bias of the dataset, ModelNet40. Hence, rotational augmentation is essential for multi-view evaluation. However, RSMix improves discriminative ability of the model with appropriate combinations with other augmentations regardless of evaluation type, because diversity of datasets increases significantly when RSMix is used in conjunction with rotation and scaling augmentations. \\
\indent In addition, we also provide the results for DGCNN~\cite{wang2019dynamic} in \tablename~\ref{table:dgcnn_ablation} with single-view evaluation on ModelNet40 and ModelNet10 with scaling augmentation as ConvDA, since single-view evaluation shows better results without rotational augmentation. We also obtained remarkable improvements with RSMix for all presented combinations.\\
\indent Therefore, we can notice three things as follows.
\begin{itemize}
	\item Single and multi-view evaluation performances differ significantly depending on the presence of rotational biases in the dataset.
	\item Rotational augmentation reduces single-view evaluation performance, but must be included when training if evaluation is performed with multi-view evaluation.
	\item RSMix successfully improved model generality by appropriate combination with other augmentation strategies regardless of evaluation type.
\end{itemize}
\begin{table}
   \tiny
      \resizebox{\columnwidth}{!}{
         \begin{tabular}{ccccc}
	        \midrule
            ConvDA & RandDrop  & RSMix & ACC\(_{S}\)(\%) & Dataset \\
	        \specialrule{0.3pt}{1pt}{0.7pt}
	        \specialrule{0.3pt}{0.7pt}{1pt}
            &  & & 92.5 & MN40  \\
            &  & \ding{51} & 93.3(0.8\(\uparrow\)) & MN40  \\
            \ding{51} & &  & 92.6 & MN40 \\
            \ding{51} & & \ding{51} & 93.4(0.8\(\uparrow\)) & MN40 \\
            \ding{51} & \ding{51} &  &  92.8 & MN40 \\
            \ding{51} & \ding{51} & \ding{51} &  \textbf{93.5(0.7\(\uparrow\))} & MN40 \\
	        \specialrule{0.2pt}{1pt}{1pt}
            &  & & 94.6 & MN10  \\
            &  & \ding{51} & 95.9(1.3\(\uparrow\)) & MN10  \\
            \ding{51} &  &  & 94.8 & MN10 \\
            \ding{51} & & \ding{51} & 95.4(0.6\(\uparrow\)) & MN10 \\
            \ding{51} & \ding{51} &  &  94.8 & MN10 \\
            \ding{51} & \ding{51} & \ding{51} &  \textbf{95.5(0.7\(\uparrow\))} & MN10 \\
            \midrule
      \end{tabular}}
   \caption{Ablation studies for DGCNN\cite{wang2019dynamic} on ModelNet40(MN40) and ModelNet10(MN10). Random scaling augmentation was applied as ConvDA.}
   \vspace{-0.5cm}
	\label{table:dgcnn_ablation}
\end{table}
\begin{table}[t!]
      \resizebox{\columnwidth}{!}{
         \begin{tabular}{c|cc|cc|c}
	        \specialrule{1pt}{1pt}{1pt}
            \multirow{2}{*}{Transform} & \multirow{2}{*}{\(\times\)} & \multirow{2}{*}{RSMix} & ConvDA  & ConvDA & \multirow{2}{*}{Eval}\\
            & & & w/o RSMix & w/ RSMIx & \\
	        \specialrule{0.4pt}{1pt}{1pt}
	        \specialrule{0.4pt}{1pt}{1pt}
			\multirow{2}{*}{Jitter (\(\sigma^{2}\)=0.01)} & 90.0 & \textbf{90.6} & 90.8 & \textbf{91.0} & Single \\
			& \textbf{82.3} & 78.5 & 90.9 & \textbf{91.3} & Multi \\
	        \specialrule{0.4pt}{1pt}{1pt}
			\multirow{2}{*}{Jitter (\(\sigma^{2}\)=0.05)} & \textbf{15.3} & 13.0 & \textbf{23.4} & 21.2 & Single \\
			& \textbf{10.3} & 9.8 & \textbf{23.5} & 20.3 & Multi \\
	        \specialrule{0.4pt}{1pt}{1pt}
			\multirow{2}{*}{X-axis \(90^{\circ}\)} & 17.3 & \textbf{21.8} & 16.5 & \textbf{19.9} & Single \\
			& 18.0 & \textbf{22.1} & 16.7 & \textbf{19.8} & Multi \\
	        \specialrule{0.4pt}{1pt}{1pt}
			\multirow{2}{*}{Y-axis \(90^{\circ}\)} & 56.7 & \textbf{60.9} & 90.0 & \textbf{91.0} & Single \\
			& 57.0 & \textbf{61.1} & 90.1 & \textbf{91.1} & Multi \\
	        \specialrule{0.4pt}{1pt}{1pt}
			\multirow{2}{*}{Z-axis \(90^{\circ}\)} & 14.6 & \textbf{18.9} & 15.2  & \textbf{18.3} & Single \\
			& 14.9 & \textbf{19.3} & 14.8 & \textbf{18.4} & Multi \\
	        \specialrule{0.4pt}{1pt}{1pt}
			\multirow{2}{*}{X-axis \(180^{\circ}\)} & 44.6 & \textbf{50.8} & 43.3 & \textbf{45.7} & Single \\
			& 44.9 & \textbf{51.1} & 43.5 & \textbf{45.6} & Multi \\
	        \specialrule{0.4pt}{1pt}{1pt}
			\multirow{2}{*}{Y-axis \(180^{\circ}\)} & 75.5 & \textbf{79.2} & 90.2 & \textbf{91.0} & Single \\
			& 75.0 & \textbf{79.1} & 90.1  & \textbf{91.1} & Multi \\
	        \specialrule{0.4pt}{1pt}{1pt}
			\multirow{2}{*}{Z-axis \(180^{\circ}\)} & 42.3 & \textbf{47.7} & 43.8 & \textbf{44.4} & Single \\
			& 41.7 & \textbf{47.9} & 43.3  & \textbf{44.4}  & Multi \\
	        \specialrule{0.4pt}{1pt}{1pt}
			\multirow{2}{*}{Scale (0.6)} & 90.7 & \textbf{92.2} & 90.4 & \textbf{91.0} & Single \\
			& \textbf{83.7} & 82.5 & 90.5 & \textbf{91.5} & Multi \\
	        \specialrule{0.4pt}{1pt}{1pt}
			\multirow{2}{*}{Scale (1.4)} & 90.7 & \textbf{92.1} & 90.5 & \textbf{91.2} & Single \\
			& \textbf{83.4} & 82.7 & 90.6 & \textbf{91.4} & Multi \\
	        \specialrule{0.4pt}{1pt}{1pt}
			\multirow{2}{*}{Scale (2.0)} & 90.7 & \textbf{92.3} & 90.4 & \textbf{91.0} & Single \\
			& \textbf{83.6} & 82.6 & 90.5 & \textbf{91.1} & Multi \\
	        \specialrule{0.4pt}{1pt}{1pt}
			\multirow{2}{*}{DropPoint (0.2)} & 88.2 & \textbf{91.9} & 87.4 & \textbf{91.0} & Single \\
			& 77.1 & \textbf{81.9} & 88.2 & \textbf{90.9} & Multi \\
	        \specialrule{1pt}{1pt}{1pt}
      \end{tabular}}
   \caption{Robustness test of RSMix with or without ConvDA for PointNet++~\cite{qi2017pointnet++} on ModelNet40 using Random shift, scaling, rotation, and jitter augmentations as ConvDA.}
   \vspace{-0.5cm}
   \label{table:Robustness_test}
\end{table}
\noindent \textbf{Robustness Test.} We tested the robustness of RSMix with PointNet++~\cite{qi2017pointnet++} to four noisy environments: jitter, rotation, scaling, and DropPoint, in order to verify that our method makes the model robust to noise. \tablename~\ref{table:Robustness_test} verifies the impact of RSMix with single-view and multi-view evaluation for 2 cases against the use of the ConvDA. Especially, multi-view evaluation for rotational noisy environment along the y-axis was performed by rotating the sample along the x-axis for fair comparison. ConvDA includes jittering, shifting, scaling, and rotational augmentations with default settings as same as PointNet++~\cite{qi2017pointnet++}. The results in \tablename~\ref{table:Robustness_test} reveal that RSMix improves the robustness of model whether or not ConvDA was applied for rotation and DropPoint noise, since shape and scale of original point cloud were preserved. However, we achieved lower results with multi-view evaluation when only RSMix was applied for scaling noise. The reason is that if scaling noise is large, it is difficult for the model to interpret the data when viewed from different angles since shape is preserved but scale compared with the original data. Meanwhile, results differed greatly depending on the level of noise for jittering noise, where the shape of an object was not preserved. Although RSMix alone cannot improve robustness for multi-view evaluation, RSMix provided improvements when jitter noise was small for single-view evaluation regardless of ConvDA usage. However, robustness was reduced for both evaluation methods for large jitter noise when RSMix was applied because it was difficult for subsets extracted from each sample by RSMix to have sufficient shape information since the original sample shape was greatly distorted prior to applying RSMix.\\[2pt]
%
\noindent \textbf{Various \(\theta\) values.} We introduced beta function Beta(\(\theta,\theta\)), to sample \(r_{rigid}\) from the beta distribution when using \(\mathcal{A}_{ball}\) in Section~\ref{subsection:mask_region_3d}. We demonstrate the experimental results for various \(\theta\) for DGCNN~\cite{wang2019dynamic} on ModelNet40~\cite{wu20153d} in \figurename~\ref{fig:alpha_test} to figure out the effect of \(\theta\) to our model using same experimental settings. Due to a property of beta function, larger \(\theta\) implies higher probability that \(r_{rigid}\) was sampled close to 0.5. However, since we set the \(n^{max}=\) half of the number of points in sample, more frequent sampling of large \(r_{rigid}\) does not have much effect. Best accuracy was achieved for \(\theta=1.0\) for all cases. Therefore, we set \(\theta=1.0\) as default for all experiments.

%

\section{Conclusion}
This paper proposes RSMix, a novel data augmentation method for point clouds, that generates virtual mixed samples from extracted subsets from each point cloud without additional learnable parameters. We extracted the subsets from samples without shape distortion by redefining the rectangular mask for images as a subset of adjacent points from a query point in 3D space. Various experiments verified that RSMix improved deep neural networks to extract discriminative feature effectively by increasing diversity of data. In addition, extensive tests demonstrated that RSMix also improved robustness of the model to various types of noise. We further analyze the two types of evaluation method for shape classification: single and multi-view, which are utilized as evaluation metrics for the overall experiments. Experiments verified explicit differences between two methods and necessity of selecting appropriate combination with various data augmentation strategies. Extensive ablation studies also verified generic effectiveness of RSMix with various combinations with existing data augmentations.\\[4pt]
%
\noindent \textbf{Acknowledgements.} 
This work was supported by the Institute for Information and Communications Technology Promotion (IITP) funded by the Korean Government (MSIP) under Grant 2016-0-00197. This research was also supported by R\&D program for Advanced Integrated-intelligence for Identification (AIID) through the National Research Foundation of KOREA(NRF) funded by Ministry of Science and ICT (NRF-2018M3E3A1057289).

\newpage

{\small
\bibliographystyle{ieee_fullname}
\bibliography{egbib}
}

\end{document}